\documentclass[10pt,twoside,journey]{IEEEtran}

\usepackage{amsmath,amssymb,amsfonts}

\usepackage{hyperref}
\usepackage{graphicx}
\usepackage{subfig}
\usepackage{booktabs}
\usepackage{array}
\usepackage{diagbox}
\usepackage{graphicx}
\usepackage{xcolor} 
\usepackage{placeins}
\usepackage{float}

\usepackage{cite}
\usepackage{textcomp}
\usepackage{url}

\usepackage{algorithm}       
\usepackage{algpseudocode}   

\usepackage{stfloats}
\usepackage{verbatim}
\usepackage{pifont} 

\begin{document}

\title{LSDM: LLM-Enhanced Spatio-temporal Diffusion Model for Service-Level Mobile Traffic Prediction}

\author{IEEE Publication Technology,~\IEEEmembership{Staff,~IEEE,}

\author{Shiyuan Zhang, Tong Li$^*$, Zhu Xiao, Hongyang Du$^*$, Kaibin Huang
\IEEEcompsocitemizethanks
{\IEEEcompsocthanksitem T. Li and Z. Xiao are with the College of Computer Science and Electronic Engineering, Hunan University, Changsha, 410082, China. E-mail: t.li@connect.ust.hk, zhxiao@hnu.edu.cn.
\IEEEcompsocthanksitem S. Zhang, H. Du and K. Huang are with the Department of Electrical and Electronic Engineering, University of Hong Kong, Pok Fu Lam, Hong Kong SAR, China. E-mail: duhy@eee.hku.hk, huangkb@eee.hku.hk
\IEEEcompsocthanksitem *Corresponding authors
}
}
}

\maketitle

\begin{abstract}
Service-level mobile traffic prediction for individual users is essential for network efficiency and quality of service enhancement. However, current prediction methods are limited in their adaptability across different urban environments and produce inaccurate results due to the high uncertainty in personal traffic patterns, the lack of detailed environmental context, and the complex dependencies among different network services. These challenges demand advanced modeling techniques that can capture dynamic traffic distributions and rich environmental features. Inspired by the recent success of diffusion models in distribution modeling and Large Language Models (LLMs) in contextual understanding, we propose an LLM-Enhanced Spatio-temporal Diffusion Model (LSDM). LSDM integrates the generative power of diffusion models with the adaptive learning capabilities of transformers, augmented by the ability to capture multimodal environmental information for modeling service-level patterns and dynamics. Extensive evaluations on real-world service-level datasets demonstrate that the model excels in traffic usage predictions, showing outstanding generalization and adaptability. After incorporating contextual information via LLM, the performance improves by at least $2.83\%$ in terms of the coefficient of determination. Compared to models of a similar type, such as CSDI, the root mean squared error can be reduced by at least $8.29\%$. The code and dataset will be available at:  \href{https://github.com/SoftYuaneR/LSDM}{https://github.com/SoftYuaneR/LSDM}.

\end{abstract}

\begin{IEEEkeywords}
Large language model, diffusion model, mobile traffic, spatio-temporal prediction
\end{IEEEkeywords}

\section{Introduction}
Mobile traffic prediction is a vital task in mobile networks, which is key in optimizing network performance and ensuring seamless connectivity for users~\cite{wu2021deep, gong2024kgda}. Accurate predictions allow network providers to allocate resources proactively and efficiently, reducing latency and preventing congestion during peak usage~\cite{huang2023safe, sun2024dynamic}. Moreover, it enhances energy efficiency by enabling operators to adjust the working status of base stations, minimizing unnecessary power consumption dynamically~\cite{li2023carbon, li2023artificial}. As networks evolve towards 6G, understanding and predicting traffic at the service level becomes essential to support emerging applications with diverse quality of service requirements, from extended reality to holographic communications. This emerging demand for fine-grained prediction is driven by the increasing heterogeneity of mobile services, where each application demands distinct network resources and user behavior patterns exhibit complex dynamics.

Significant effort has been devoted to mobile traffic prediction in recent years. Initially, researchers approached this task as a general time-series forecasting problem, employing techniques such as Recurrent Neural Networks (RNNs) and Long Short-Term Memory (LSTM) networks to model the temporal dependencies of traffic at individual base stations~\cite{trinh2018mobile, feng2018deeptp}. Furthermore, subsequent studies acknowledged the importance of spatial correlations between base stations and incorporated Graph Neural Networks (GNNs) into their frameworks, leveraging graph-based structures to model the spatial distribution among base stations~\cite{wang2022spatial, gong2023empowering}.
While these approaches have achieved notable progress in predicting aggregated traffic at the base station level, they cannot effectively model the complex and dynamic behaviors of individual mobile users~\cite{fang2022sdgnet}. Specifically, the lack of user-level granularity limits their applicability in enabling advanced network optimizations, such as fine-tuned resource allocation and personalized Quality-of-Service (QoS) guarantees. To support more precise and proactive traffic management, there is a pressing need for models that capture the individualized and service-level traffic usage behavior of mobile users, facilitating a shift from aggregated analysis to user-centric and service-specific network optimization strategies.

However, developing such service-level mobile traffic prediction models for individual users presents several fundamental challenges. 
\textbf{Firstly}, unlike aggregated traffic usage at base stations, mobile users' traffic usage exhibits significant fluctuations and high uncertainty due to the variability in individual user behaviors. This challenge is particularly pronounced at the service level, where usage patterns vary widely depending on user preferences, habits, and service demands. Capturing these highly dynamic and service-specific traffic patterns across different users and accurately modeling the fluctuations is a critical and complex issue.
\textbf{Secondly}, mobile users' behaviors are heavily influenced by environmental factors. In other words, users access different network services depending on the functional region they are in, such as residential, commercial, or public areas. For instance, users in residential areas may favor streaming services like Netflix or YouTube, while users in commercial zones might primarily rely on productivity tools or communication services such as Zoom or Microsoft Teams~\cite{li2020extent, yang2023atpp}. Accurately extracting these environmental factors and understanding their impact on service-level usage behaviors is crucial for creating an effective model.
\textbf{Thirdly}, different network services are often used in combination, leading to high interdependence among their usage patterns. For example, while streaming a live sports event on YouTube, users may simultaneously use messaging apps like WhatsApp to chat with friends or check social media platforms like Twitter to follow game updates~\cite{liao2013mining}. These compound usage patterns create intricate dependencies between services, which must be carefully captured and modeled to ensure accurate service-level traffic predictions.

To address these challenges, we propose a Large Language Model (LLM)-enhanced Spatiotemporal Diffusion Model (LSDM) for service-level mobile traffic prediction for individual mobile users. Specifically, LSDM leverages a conditional diffusion model as the backbone for traffic prediction. Unlike traditional deterministic models such as LSTMs and RNNs~\cite{sherstinsky2020fundamentals}, diffusion models are probabilistic and excel at modeling the joint distribution of complex data~\cite{croitoru2023diffusion}. This capability allows LSDM to better account for the uncertainty and variability inherent in individual user traffic usage behaviors. By gradually denoising data and conditioning on impactful environmental features, the LSDM framework effectively models mobile traffic fluctuations, addressing the \textbf{first} challenge of capturing highly dynamic and uncertain traffic usage patterns.
Furthermore, LSDM introduces an innovative approach to incorporate the environmental features of mobile users by leveraging easily accessible satellite images. Compared to traditional methods that rely on coarse-grained geographical information, such as region-level demographics or static region classifications, satellite images provide abundant texture information about the physical environment, including building shapes, spatial distributions, road networks, and vegetation patterns. LSDM further enhances these visual features through LLM-generated textual descriptions corresponding to the satellite images. For instance, a satellite image of a commercial district might generate a description such as \textit{``area with highdensity
office buildings, retail shops, and public transportation
access.''} To integrate these multimodal features effectively, we design a multimodal contrastive learning method that integrates satellite image features, LLM-generated textual descriptions, and Points of Interest (POI)~\cite{ye2011exploiting} features (e.g., nearby landmarks, businesses, or facilities), creating comprehensive environmental representations that serve as conditional inputs of the diffusion model for traffic prediction. This enables the model to capture and understand the impact of environmental factors on service-level traffic usage behaviors, thereby addressing the \textbf{second} challenge. To address the \textbf{third} challenge, capturing the interdependence among different network services, LSDM introduces a two-dimensional attention mechanism based on a double-layer Transformer architecture~\cite{zhang2024netdiff}. The first attention layer is designed to model the temporal dynamics of traffic usage at the individual service level, capturing patterns and fluctuations over time. The second layer focuses on feature correlations between different network services, effectively modeling the interdependencies in service usage. This dual-layer attention structure enables comprehensive modeling of the complex relationships between various apps and their traffic patterns, leading to accurate and reliable traffic predictions.

The contributions of this paper are summarized as follows:
\begin{itemize}
    \item We clearly define the task of user- and service-level mobile traffic prediction, addressing key challenges such as high dynamics, uncertainty, environmental impact, and service interdependencies.
    \item We propose the LSDM framework, which leverages a conditional diffusion model enhanced by LLMs for multimodal environmental feature modeling, combining satellite imagery, LLM-generated textual descriptions, and POI data. Additionally, LSDM employs a dual-layer Transformer to capture temporal dynamics and service interdependencies.
    \item  We validate LSDM through extensive testing on real-world datasets, demonstrating its superior performance over existing methods. After incorporating contextual information, the coefficient of determination improves by at least $2.83\%$, and compared to models of a similar type, the Root Mean Squared Error (RMSE) can be reduced by at least $8.29\%$.

\end{itemize}

The rest of this paper is organized as follows. In Section~\ref{related_work}, we present the related work on spatiotemporal prediction and conditional diffusion models in this field, and the use of large models to handle multimodal data. In Section~\ref{pre_and_obj}, we provide a detailed definition of mobile user usage prediction, the principles of diffusion models in prediction tasks, and our task objectives. We then provide a detailed explanation of the structure of LSDM and the design details in Section~\ref{lsdm}. In Section~\ref{per}, we provide a detailed description of the experimental setup and the performance evaluation of the model. In Section~\ref{ab}, we design relevant ablation experiments to further compare and analyze the different modules of LSDM.

\section{Related Work}
\label{related_work}
In this section, we discuss several related works, including spatio-temporal prediction, conditional diffusion models, and LLM for multimodal representations.

\subsection{Mobile Traffic Prediction}

Mobile traffic prediction aims to forecast future network traffic patterns by analyzing historical data on user behavior, application usage, and network resource consumption in mobile communication systems~\cite{jiang2022cellular, jiang2024mobile}. The rise of deep learning techniques has significantly advanced mobile traffic prediction. A diverse range of models has been developed to capture spatio-temporal relationships, encompassing CNN-based~\cite{li2017diffusion, liu2018attentive, zhang2017deep}, RNN-based~\cite{feng2018deeptp, wang2018predrnn++, wang2017predrnn}, GNN-based~\cite{jiang2024mobile, he2020graph}, Transformer-based~\cite{wang2024transformer, yu2020spatio, shuvro2023transformer}, MLP-based~\cite{nikravesh2016mobile}, GAN-based~\cite{wu2021deep}, probabilistic models based~\cite{aceto2021characterization}, and diffusion model-based architectures~\cite{yuan2023spatio, zhou2023towards, andreoletti2019network, chai2025spatio}.
Simultaneously, cutting-edge learning techniques like meta-learning, contrastive learning, and adversarial learning are also employed~\cite{wu2021deep, yin2022practical}. However, most existing methods in mobile traffic prediction struggle with adaptability in handling dynamic network traffic patterns, as user behavior and application usage characteristics may change significantly over time, leading to a decline in prediction accuracy~\cite{li2024mobile, sun2021mobile}. Existing approaches in such dynamic scenarios often rely on overly complex mechanisms to model contextual and conditional information. This added complexity frequently hinders generalization, reducing their effectiveness in adapting to diverse or unforeseen network conditions. In contrast, our model enables accurate traffic predictions in these dynamic scenarios and incorporates contextual information in a more efficient and adaptable manner.


\subsection{Conditional Diffusion Models for Spatio-temporal Data}

Inspired by the remarkable strides in conditional diffusion models for image and data generation~\cite{batzolis2021conditional, luo2024scdiffusion}, the exploration of conditional diffusion model for mobile traffic has emerged as a compelling problem.

Diffusion models have emerged as a more effective alternative to GANs for data generation due to their simpler training process. Unlike GANs, which require the interaction of a generator and a discriminator—often leading to issues like mode collapse—diffusion models utilize a noise-mixing and denoising process~\cite{zhang2024netdiff}. This approach not only stabilizes training but also enables the generation of high-quality data.

A foundational innovation in this field is the introduction of Denoising Diffusion Probabilistic Models (DDPM), which set the stage for subsequent advancements. Building on this, conditional diffusion models, such as those enhanced with additional conditioning steps, allow for more diverse and detailed generation tailored to specific scenarios. For example, advanced diffusion techniques have demonstrated the capability to produce outputs with highly specific attributes by employing layered or conditional approaches, showcasing their versatility in various domains.

While existing methods in network trace generation have achieved some success, they predominantly focus on generating data for single network services~\cite{sivaroopan2024netdiffus, dowoo2019pcapgan}. This narrow scope fails to capture the diversity and complexity of interactions across multiple applications, limiting their effectiveness in real-world scenarios. Furthermore, GAN-based approaches commonly used in this context are prone to mode collapse, making them less reliable for practical applications. In contrast, diffusion models have been proven to generate high-fidelity data and overcome such limitations, offering a robust solution for simulating complex network environments. These models not only address the challenges of representing diverse service types but also provide a more realistic reflection of user interactions and network dynamics.

\subsection{Large Language Models for Multimodal Representations}
With the rapid advancements in LLMs, their applications have expanded beyond text-only tasks to include multimodal integration, where textual, visual, and other sensory inputs are processed collaboratively~\cite{minaee2024large}. LLM-driven multimodal integration leverages the semantic understanding and reasoning capabilities of LLMs to unify heterogeneous data modalities into a coherent representation. This approach has enabled various downstream tasks, such as cross-modal retrieval, multimodal question answering, and content generation~\cite{minaee2024large}.

CLIP introduced a groundbreaking approach by training a dual-tower architecture to contrastively align image and text representations from large-scale Internet datasets~\cite{hafner2021clip}. Subsequent developments in CLIP models have focused on scaling both model and dataset sizes, introducing self-supervised techniques, improving the efficiency of pre-training processes, and enabling few-shot learning capabilities. While CLIP models are predominantly trained on general image datasets, there has been increasing interest in adapting them to specific domains. In healthcare, models like ConVIRT~\cite{zhang2022contrastive}, PubMedCLIP~\cite{eslami2023pubmedclip}, MedCLIP~\cite{wang2022medclip}, and BioMedCLIP~\cite{zhang2023biomedclip} have achieved significant progress in medical imaging applications. Similarly, specialized CLIP variants trained on large-scale E-commerce datasets have demonstrated notable improvements over baseline models. Beyond these efforts, CLIP has also been employed in creative domains, powering applications in text-to-image generation and multimodal understanding~\cite{liu2024remoteclip}. Despite these advancements, the use of CLIP in remote sensing remains underexplored. While Zhang et al. have initiated efforts by utilizing aerial imagery from large-scale datasets to train CLIP models, opportunities abound for adapting these models to tasks in remote sensing, such as land cover analysis, disaster assessment, and environmental monitoring.

\section{Preliminaries and Objective}
\label{pre_and_obj}
\begin{figure}[t] 
    \centering
    \includegraphics[width=\columnwidth]{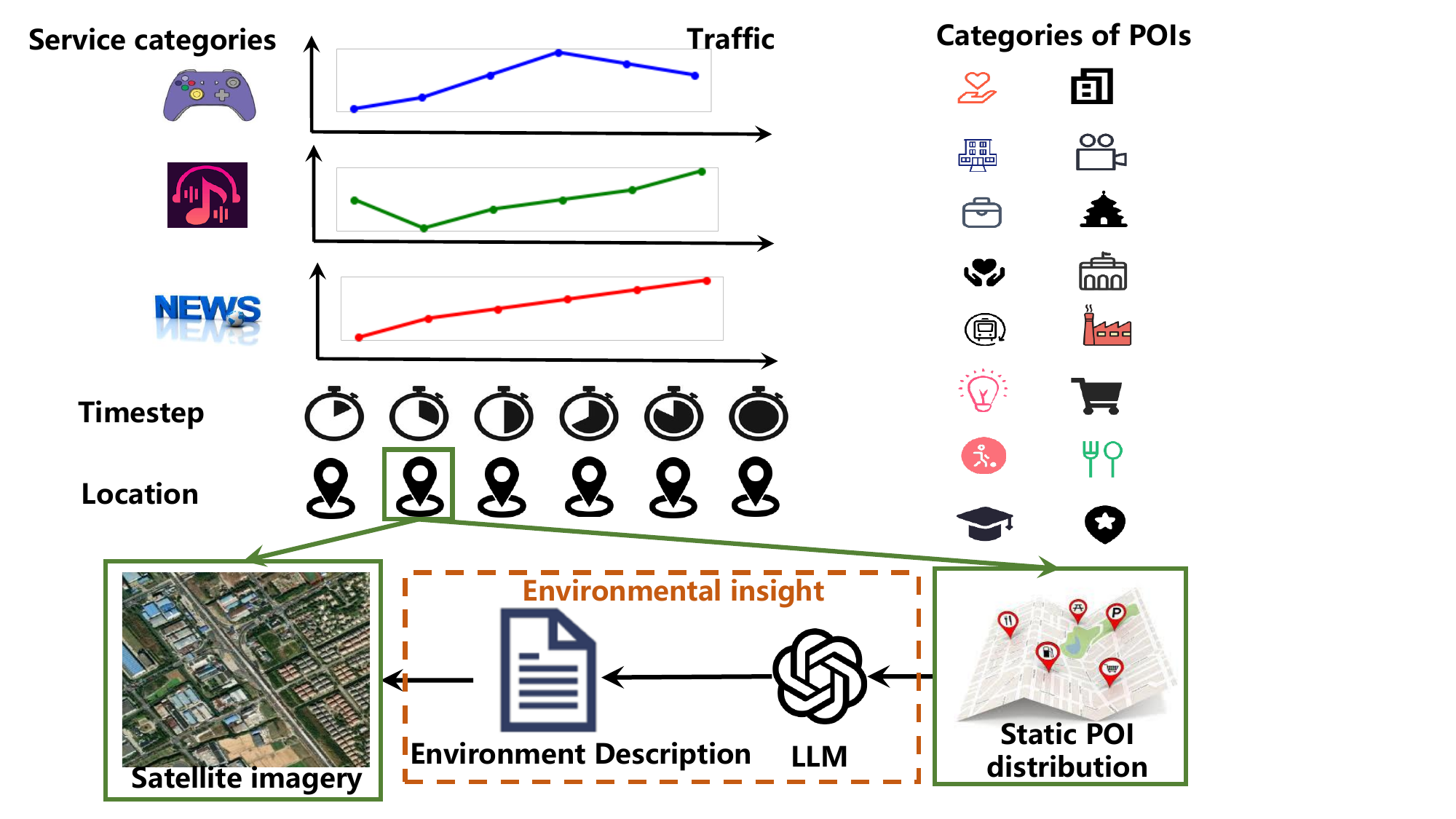} 
    \caption{Overview of Data Format. The data shows the traffic generated by a user utilizing different services, with the user's location available at each time point. The corresponding environmental information includes a satellite image of that location and the number of POIs. We utilize an LLM to transform the POI count information into descriptive text for the image.} 
    \label{fig_data} 
\end{figure}

In this section, we define the problem of mobile user usage prediction, explain the principles of diffusion models in prediction tasks, and outline the objectives that guide our approach.
\subsection{Spatio-temporal Mobile Traffic Usage Data}
Each user's app usage is modeled as a two-dimensional matrix $\mathbf{A}^{M \times N}$, where the corresponding spatial point information is represented by $\mathbf{B}^{M \times P}$, where $P$ denotes the number of POI categories surrounding the spatial point at each sampling time, let the satellite image data corresponding to each sampling time be represented by $\mathbf{S}^{M \times H \times W \times C}$, where $H$, $W$, and $C$ denote the height, width, and the number of channels of the satellite images, respectively.

\begin{itemize}
    \item $M = 168$, representing 168 sampling points over one week, with a time granularity of one hour;
    \item $N = 10$, representing 10 app categories (services);
    \item Each element $A_{i,j}$ of the matrix $\mathbf{A}$ denotes the traffic generated by the user using the $j$-th app category during the $i$-th hour;
    \item $P = 17$, representing the types of POIs (Points of Interest) surrounding the user's geographic location at the sampling time.
\end{itemize}

The specific data examples are shown in Figure~\ref{fig_data}. It is worth mentioning that the framework is not confined to the specific settings used in the experiments. The framework itself is highly general and can accommodate different application scenarios and requirements. The parameters used in the experiments, such as time granularity, service categories, and POI types, were selected based on specific experimental settings. These are not limitations of the framework but rather reference values to validate its performance. The framework is designed to dynamically adjust and provide adaptable solutions to meet the demands of various tasks.

\begin{table}[tb]
\centering
\caption{App/service Categories and Descriptions}
\label{tab:app_categories}
\renewcommand\arraystretch{1.25}
\begin{tabular}{c|l|m{4.5cm}}
\toprule
\textbf{Index} & \textbf{App Category}      & \textbf{Examples/Description}                   \\ \hline
1              & Utilities                  & General utilities for daily use                \\ \hline
2              & Games                      & Gaming applications                             \\ \hline
3              & Entertainment              & Streaming platforms, video apps                \\ \hline
4              & News                       & News applications for current affairs          \\ \hline
5              & Social Networking          & Apps for social interactions like WeChat, LinkedIn, Weibo \\ \hline

6              & Travel                     & Travel planning and booking applications       \\ \hline
7              & Lifestyle                  & Lifestyle-related apps like Meituan            \\ \hline
8             & Navigation                 & Navigation and GPS tools                       \\ \hline
9             & Music                      & Music streaming platforms                      \\ \hline
10             & Photo \& Video             & Photography and video editing apps             \\ 
\bottomrule
\end{tabular}
\end{table}

\begin{table}[tb]
\centering
\caption{Categories of Points of Interest (POIs)}
\label{tab:poi_categories}
\renewcommand\arraystretch{1.2}
\begin{tabular}{c|l|c|l|c|l}
\toprule
\multicolumn{6}{c}{\textbf{POI Categories}} \\
\hline
\textbf{ID} & \textbf{Category} & \textbf{ID} & \textbf{Category} & \textbf{ID} & \textbf{Category} \\
\hline
1 & Medical Care     & 7  & Sports            & 13 & Shopping        \\\hline
2 & Hotel           & 8  & Residence         & 14 & Restaurant      \\\hline
3 & Business        & 9  & Entertainment     & 15 & Education       \\\hline
4 & Life Service    & 10 & Scenic Spot       & 16 & Landmark        \\\hline
5 & Transport Hub   & 11 & Government        & 17 & Other           \\\hline
6 & Culture         & 12 & Factory           & \multicolumn{2}{c}{\diagbox{}{}} \\
\bottomrule
\end{tabular}
\end{table}

\subsection{Diffusion Models}  
\label{sec:diff}  
DDPMs are a class of generative models that operate by reversing a structured noise-addition process~\cite{yang2023diffusion}. Specifically, DDPMs work through a forward process that gradually adds noise to data samples and a learned reverse process that removes this noise step by step. By training the model to reverse the noise addition, DDPMs learn to generate new samples that match the original data distribution. This iterative denoising mechanism allows the model to capture intricate patterns and relationships within the data, making it particularly well-suited for complex generative tasks.

\textbf{Diffusion Process (Noise Addition):}  
The forward diffusion process involves gradually introducing noise into the data through a series of Markov steps. Starting from an initial data point $x_0$, this process transforms it step-by-step into a fully noised representation $x_T$. Each step can be expressed as:  
\begin{equation}  
x_t = \sqrt{\alpha_t} x_{t-1} + \sqrt{1 - \alpha_t} \epsilon,  
\end{equation}  
where $\alpha_t$ is a predefined parameter controlling the noise scale at step $t$, and $\epsilon$ is sampled from a standard normal distribution. This structured noise addition results in a predictable trajectory from data to pure noise.

\textbf{Reverse Diffusion (Noise Removal):}  
To generate data, the reverse process undoes the added noise step by step, transforming noise back into structured data. This reverse process is also a Markov chain, where a neural network predicts the noise at each step and refines the intermediate representation. The reverse step is mathematically defined as:  
\begin{equation}  
x_{t-1} = \frac{1}{\sqrt{\alpha_t}} \left( x_t - \frac{1 - \alpha_t}{\sqrt{1 - \bar{\alpha}_t}} \epsilon_\theta(x_t, t) \right),  
\end{equation}  
where $\epsilon_\theta(x_t, t)$ denotes the predicted noise by the model at step $t$, and $\bar{\alpha}_t$ represents the cumulative product of $\alpha_t$ values up to step $t$. Training the model involves minimizing the discrepancy between the predicted and actual noise, using the following loss function:  
\begin{equation}  
    \underset{\theta}{\min}\mathcal{L}(\theta):=\mathbb{E}_{{x}_0,{\bf \epsilon},t}\left[||{\bf \epsilon}-{\bf \epsilon}_\theta({x}_t,t)||^2\right].  
\end{equation}  

By systematically combining noise addition and removal processes, diffusion models offer a flexible and powerful framework for generating high-quality samples. They excel in maintaining stability during training and avoiding common pitfalls such as mode collapse, which are often encountered in alternative generative frameworks like GANs.

\label{sec:diff_pred}

Although diffusion models are primarily designed for generative tasks, their flexible framework can be adapted for prediction problems. In such cases, the objective is to infer unknown target values based on partially observed inputs. This reframes the prediction task as a conditional generation problem, where the known inputs serve as conditioning information, and the model predicts the desired outcomes.

\textbf{Modeling the Prediction Process:}  
To adapt diffusion models for prediction, the forward diffusion process remains unchanged, where noise is gradually added to the data. However, the reverse process is conditioned on observed inputs $c$, such that the model learns to infer the denoised target data $x_0$ given both the noisy data $x_t$ and the condition $c$. Mathematically, the reverse step is expressed as:  
\begin{equation}  
x_{t-1} = \frac{1}{\sqrt{\alpha_t}} \left( x_t - \frac{1 - \alpha_t}{\sqrt{1 - \bar{\alpha}_t}} \epsilon_\theta(x_t, t, c) \right),  
\end{equation}  
where $\epsilon_\theta(x_t, t, c)$ is the predicted noise by the model, conditioned on $c$.

\textbf{Training Objective:}  
To train the model for prediction, the loss function is modified to incorporate the conditioning information, ensuring the model effectively utilizes the observed inputs. The training objective becomes:  
\begin{equation}  
    \underset{\theta}{\min} \mathcal{L}(\theta):= \mathbb{E}_{{x}_0, c, {\bf \epsilon}, t}\left[||{\bf \epsilon}-{\bf \epsilon}_\theta({x}_t, t, c)||^2\right].  
\end{equation}  

\textbf{Applications in Prediction:}  
Diffusion models for prediction excel in tasks requiring high-dimensional or structured data inference, such as time-series forecasting, image inpainting with missing regions, and trajectory prediction. By leveraging their iterative denoising approach and ability to model complex data distributions, diffusion-based prediction models can outperform traditional regression or sequence modeling techniques, especially when handling multimodal uncertainties or intricate dependencies in the data.

This framework highlights the adaptability of diffusion models, demonstrating their ability to address predictive tasks by conditioning the denoising process on observed data, enabling accurate and robust predictions across diverse applications.


\subsection{Mobile Traffic Usage Prediction with Diffusion Models}
For a specific dataset, given historical observations, we aim to predict the future one step. The spatio-temporal prediction task can be formulated as learning a $\theta$-parameterized model $\mathcal{F}$:
\begin{equation}
    A_{[t+1]} = \mathcal{F}_\theta(A_{[t-H:t]}, B_{[t-H:t]}, S_{[t-H:t]}),
\end{equation}
where $A_{[t+1]}$ represents the predicted future mobile traffic usage at time $t+1$, and $A_{[t-H:t]}$, $B_{[t-H:t]}$, and $S_{[t-H:t]}$ denote the historical mobile traffic data, external features, and spatial information, respectively, over the past $H$ time steps.

To address this task with diffusion models, we introduce a conditional diffusion framework. In this setup, the model learns to predict future traffic usage by reversing a noise-adding diffusion process, where noisy versions of historical data are progressively refined to obtain the prediction. The model is conditioned on both historical mobile traffic data and external features, ensuring the predictions account for temporal dependencies and contextual factors.

Instead of treating prediction as a direct generation process, the diffusion model approaches it as an iterative refinement task. Starting with noisy initial predictions, the model progressively denoises the data step by step. At each stage, it incorporates historical traffic data, external features, and spatial information to guide the refinement process, ensuring that both temporal dependencies and contextual environment factors inform the predictions.

This iterative framework leverages the strengths of diffusion models, enabling them to capture complex temporal relationships and dynamically adapt to varying traffic patterns and external influences. By effectively integrating diverse contextual information, the approach enhances the robustness and accuracy of mobile traffic usage predictions, making it well-suited for dynamic and context-sensitive forecasting tasks.

\section{LLM-Enhanced Spatio-temporal Diffusion Model}
\label{lsdm}
\begin{figure*}[htbp] 
    \centering
    \includegraphics[width=0.7\textwidth]{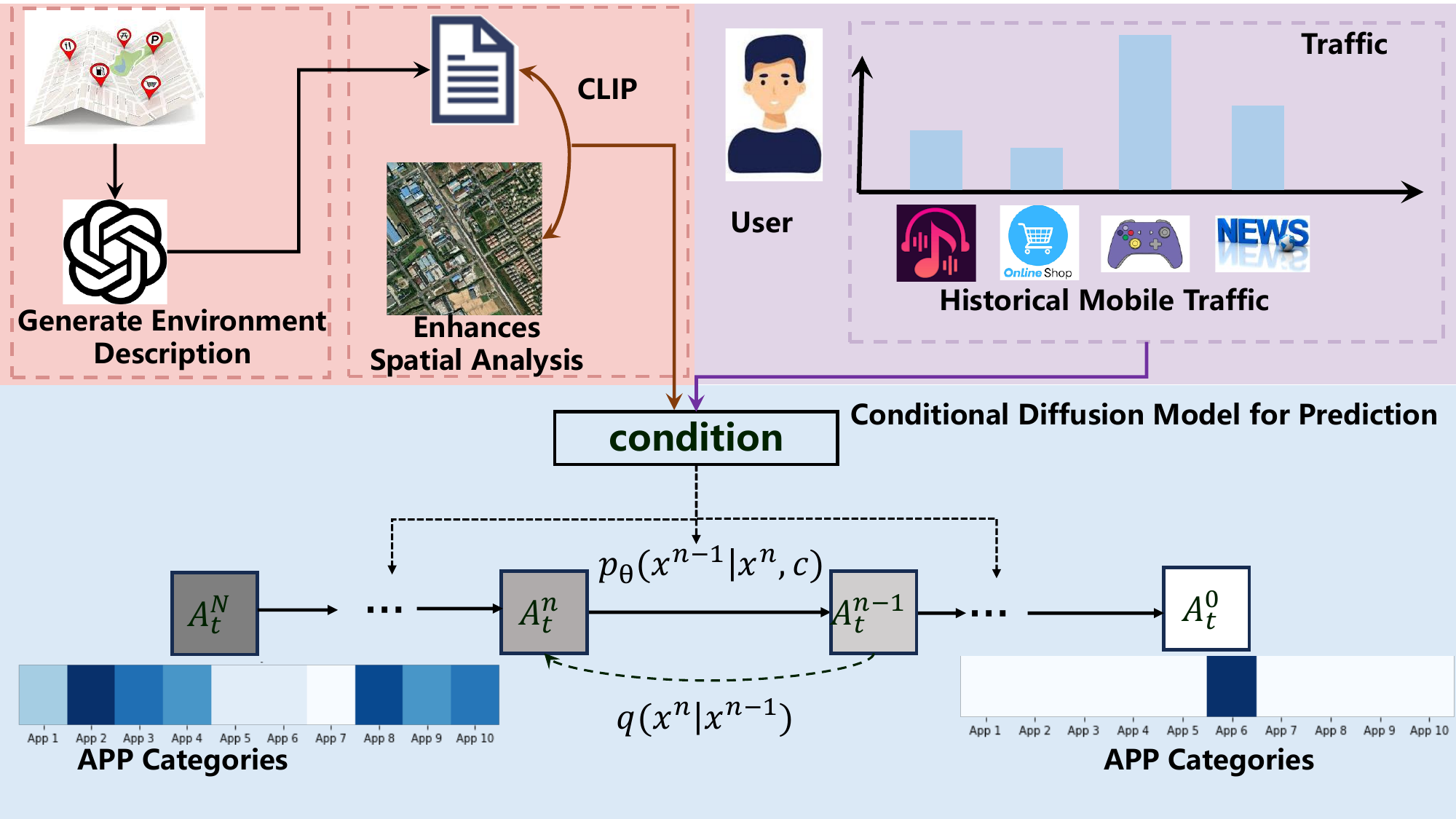} 
    \caption{Overview of LSDM. The diffusion model predicts app categories using environmental descriptions generated by the LLM, satellite imagery, and historical mobile traffic as input conditions.} 
    \label{fig_o} 
\end{figure*}

We propose the LDSM model to address spatio-temporal service-level mobile traffic usage prediction, which is shown in Figure~\ref{fig_o}. It consists of two stages.

\begin{itemize}
    \item \textbf{Stage 1: Employ LLM for processing multimodal information.} Different from existing methods that are limited to constructing specific networks for particular domain information to obtain embeddings, our approach involves collecting extensive multimodal data, including satellite images of spatial points and surrounding POI information. We leverage the background knowledge of large language models to enhance the performance of integrating multimodal information.
    \item \textbf{Stage 2: Train a conditional diffusion model for prediction.} This stage incorporates multimodal capabilities from LLMs and historical mobile traffic usage for conditional prediction, leveraging a conditional diffusion model. We design a 2D attention mechanism to capture relationships across different services.
  
\end{itemize}

\subsection{LLM-Driven Multimodal Integration}

\begin{figure}[t] 
    \centering
    \includegraphics[width=\columnwidth]{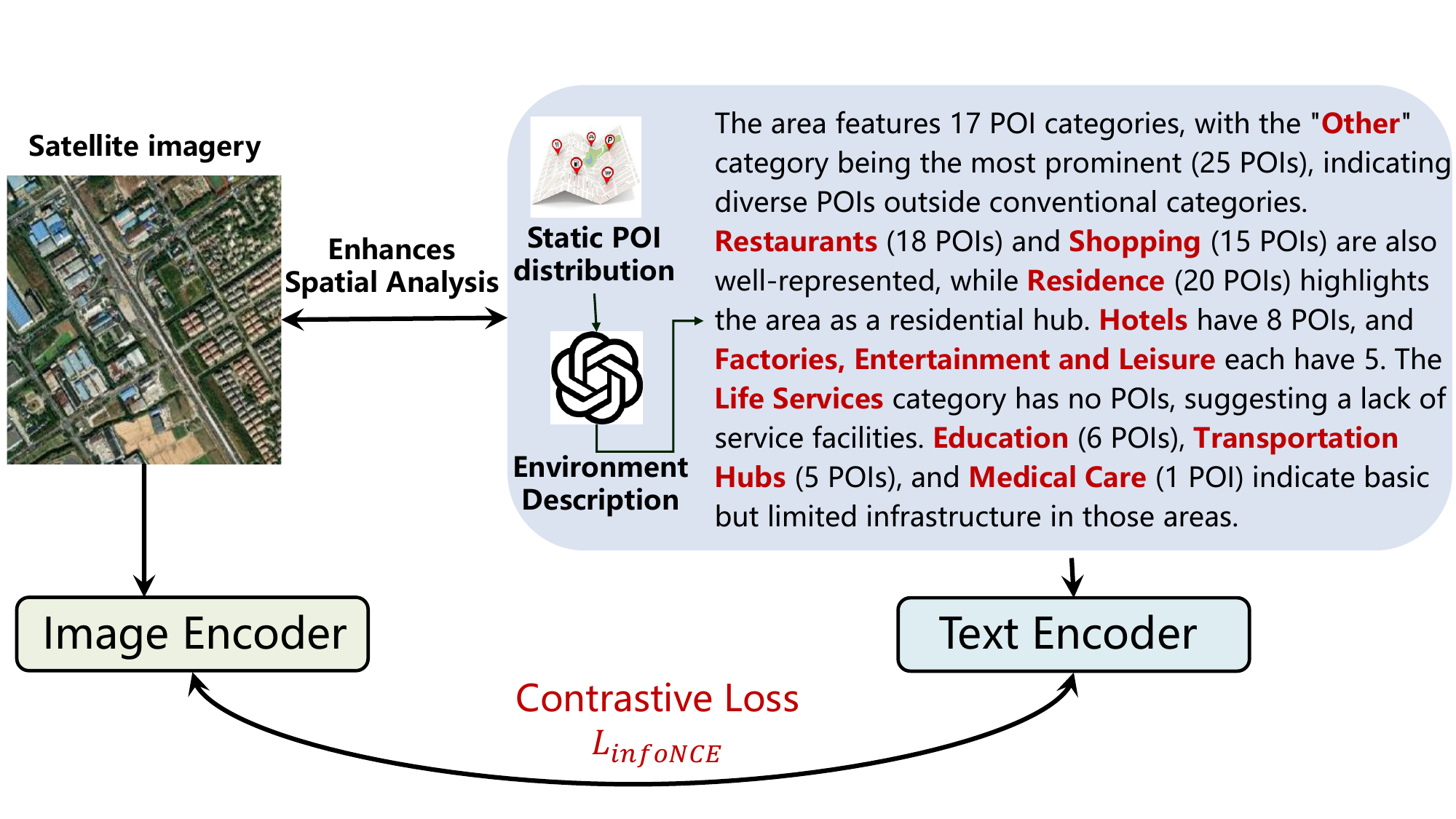} 
    \caption{Environmental insight. The numerical information of POIs is converted into textual descriptions to enhance the representation of the corresponding satellite images.} 
    \label{fig_e} 
\end{figure}

Our model leverages environmental information as input, which includes satellite images in visual form and the number of POIs in textual descriptive form. To effectively capture the embeddings of environmental information at each time step, we utilize a large model based on CLIP, which is pretrained on vast multimodal datasets and encodes rich contextual information between text and images, as illustrated in Figure~\ref{fig_e}

\subsubsection{CLIP-Based Embedding Extraction}
The CLIP model employs a contrastive learning framework to align textual and visual representations by training on paired image-text data. Specifically, CLIP comprises an image encoder $f_I$ and a text encoder $f_T$, which encode an image input $x_I$ and a text input $x_T$ into their respective latent representations:
\begin{equation}
z_I = f_I(x_I) \in \mathbb{R}^{d_z \times 1}, \quad z_T = f_T(x_T) \in \mathbb{R}^{d_z \times 1},
\end{equation}
where $d_z$ denotes the dimensionality of the latent space.

During training, CLIP optimizes a bidirectional InfoNCE loss to maximize the similarity of matched image-text pairs while minimizing the similarity of unmatched pairs. The loss function is formulated as follows:
\begin{equation}
\begin{aligned}
L_{\text{InfoNCE}} = & - \frac{1}{N} \sum_{i=1}^{N} \log \frac{\exp(z_I^i \cdot z_T^i / \tau)}{\sum_{j=1}^{N} \exp(z_I^i \cdot z_T^j / \tau)} \\
& - \frac{1}{N} \sum_{i=1}^{N} \log \frac{\exp(z_T^i \cdot z_I^i / \tau)}{\sum_{j=1}^{N} \exp(z_T^i \cdot z_I^j / \tau)},
\end{aligned}
\end{equation}
where $N$ denotes the batch size, and $\tau$ is a learnable temperature parameter.

By optimizing this loss, the CLIP model achieves two key properties:
\begin{itemize}
    \item \textbf{Representation Alignment}: Paired image-text samples exhibit high similarity ($z_I^i \cdot z_T^i$ is large), while unmatched samples have low similarity ($z_I^i \cdot z_T^j$ for $i \neq j$ is small).
    \item \textbf{Representation Grouping}: Semantically similar unimodal samples are clustered together, while distinct samples are separated.
\end{itemize}

Given the strong multimodal representation capabilities of CLIP, we encode satellite images and textual descriptions using its image encoder and text encoder, respectively, obtaining latent representations $z_I$ and $z_T$. The final environmental embedding is constructed as a weighted combination of these two representations:
\begin{equation}
z_{\text{env}} = \alpha \cdot z_I + \beta \cdot z_T,
\end{equation}
where $\alpha$ and $\beta$ are tunable parameters that control the relative contributions of image and text representations.

\subsubsection{Advantages in Our Task}

In our task, integrating environmental information through a LLM like CLIP offers the following advantages:
\begin{itemize}
    \item \textbf{Rich Contextual Understanding}: The pretrained CLIP model incorporates diverse semantic relationships between textual and visual modalities, enabling it to capture nuanced environmental features that are essential for downstream decision-making tasks~\cite{hafner2021clip}.
    \item \textbf{Multimodal Fusion}: By combining satellite imagery and textual descriptions of POIs, the model effectively merges complementary modalities, improving the robustness of environmental representations~\cite{shen2021much}.
    \item \textbf{Task Generalization}: The pretrained nature of the large model allows it to adapt to varied environmental contexts without requiring extensive retraining, making it suitable for tasks involving dynamic or heterogeneous inputs~\cite{yan2024urbanclip}.
    \item \textbf{Improved Efficiency}: The alignment between image and text embeddings reduces the need for handcrafted feature engineering, streamlining the feature extraction process while maintaining high accuracy~\cite{che2023enhancing}.
\end{itemize}

The superior performance of CLIP in multimodal tasks aligns well with the requirements of our task, as it ensures both efficiency and accuracy in leveraging environmental information.

\subsection{Conditional Diffusion model for Prediction}

In this work, we propose the use of a conditional ciffusion Model for predicting app usage patterns over time. The dataset is represented as a two-dimensional matrix \( \mathbf{X} \in \mathbb{R}^{T \times C} \), where \( T \) represents the number of time steps and \( C \) denotes the number of app categories. Each element of this matrix, \( \mathbf{X}_{t,c} \), corresponds to the app usage for the \( c \)-th category at the \( t \)-th time step. The objective is to predict the future app usage matrix \( \hat{\mathbf{X}} \) at each time step, conditioned on the historical data and other contextual information, such as user behavior and environmental factors.

\begin{figure}[t] 
    \centering
    \includegraphics[width=1.0\columnwidth]{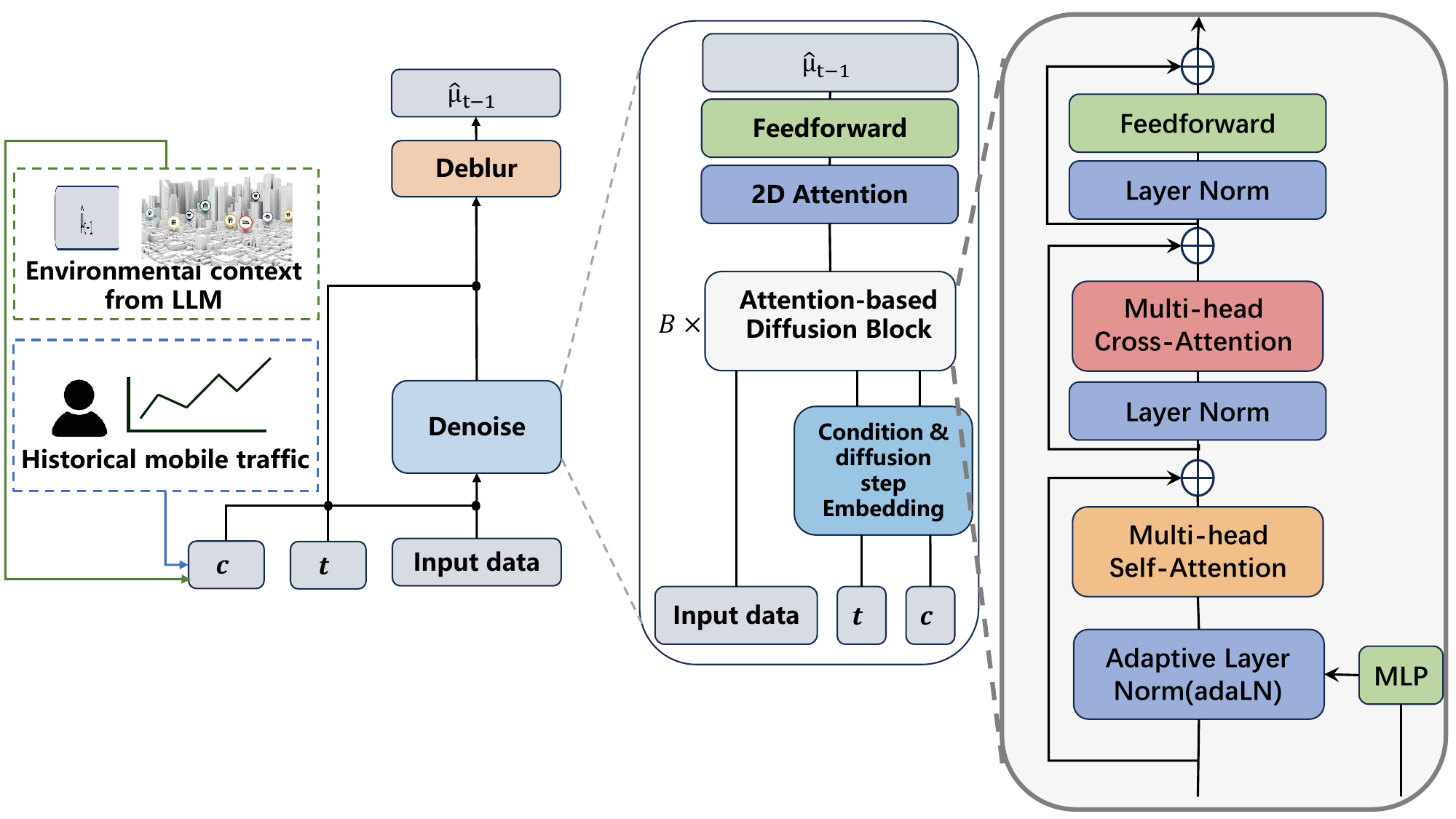}
    \caption{The structure of conditional diffusion model.} 
    \label{dit} 
\end{figure}

To achieve this, we adopt a \textbf{Conditional Diffusion Transformer (DiT)} structure, which combines the generative power of diffusion models with the expressive capability of transformers. The DiT architecture leverages a transformer backbone to model temporal dependencies and cross-category relationships, enabling it to capture intricate usage patterns across both time and app categories.The structure is shown in Figure~\ref{dit}. Specifically, the transformer-based diffusion model ensures that the historical data is effectively encoded and the temporal correlations are preserved, while the conditional mechanism integrates contextual information such as user profiles or external conditions.

The advantages of the DiT structure in this work are multifold:
\begin{enumerate}
    \item \textbf{Enhanced Temporal Dynamics Modeling}: The DiT structure, with its first-layer attention mechanism, captures fine-grained temporal variations in traffic usage at the individual service level. This enables the model to account for short-term fluctuations and long-term trends in user behavior, improving prediction accuracy~\cite{parkforecasting, grigsby2021long}.

    \item \textbf{Cross-Service Correlation Representation}: The second-layer attention mechanism effectively models the interdependencies among different network services. This allows the model to capture complex relationships and joint usage patterns between various applications, ensuring a more holistic understanding of service-level traffic behaviors~\cite{liao2024poster}.

    \item \textbf{Multimodal Feature Integration}: By incorporating both visual and textual environmental features, the DiT structure seamlessly integrates diverse data modalities. This capability enhances the model's understanding of the impact of environmental factors, such as POIs and satellite images, on user traffic patterns~\cite{ma2024multimodal, zhang2024text}.

    \item \textbf{Probabilistic Traffic Prediction}: The probabilistic nature of the diffusion model, combined with the Transformer architecture, allows the DiT structure to better handle the uncertainty and variability inherent in mobile traffic usage. This ensures robust and reliable predictions even in highly dynamic environments~\cite{feng2024latent}.

    \item \textbf{Scalability and Flexibility}: The modular design of the DiT structure supports scalability across different data scales and service types. Its flexibility allows adaptation to varying network environments and user behaviors, making it suitable for diverse traffic prediction scenarios~\cite{lu2024fit}.
\end{enumerate}

By leveraging the DiT structure, our proposed method not only achieves high prediction accuracy but also demonstrates its capacity to generalize across diverse temporal and categorical patterns, making it a powerful tool for app usage prediction tasks.

We generate predictions conditioned on the given data, i.e., the model generates a predicted matrix \( \hat{\mathbf{X}} \) based on the historical usage data \( \mathbf{H} \) and additional contextual features \( \mathbf{C} \). The diffusion process models the evolution of the data over time, where the conditioning information \( \mathbf{H} \) and \( \mathbf{C} \) guide the transitions. As illustrated in Figure~\ref{fig_o}, at each step, the model iteratively transforms the input data towards a predicted output, ensuring that the predicted usage matrix \( \hat{\mathbf{X}} \) aligns closely with the true matrix \( \mathbf{X}_{\text{true}} \).

The forward diffusion process is modeled as a series of stochastic transitions. At each time step \( t \), noise is added to the data \( \mathbf{x}_t \), conditioned on \( \mathbf{H} \) and \( \mathbf{C} \), to create the noisy observation \( \mathbf{x}_{t+1} \) as:
\[
q(\mathbf{x}_{t+1} | \mathbf{x}_t, \mathbf{H}, \mathbf{C}) = \mathcal{N}(\mathbf{x}_{t+1}; \mathbf{x}_t, \beta_t \mathbf{I}),
\]
where \( \mathbf{x}_t \) represents the data at time step \( t \), and \( \beta_t \) is a noise schedule that determines the amount of noise added. The model is trained to reverse this diffusion process and recover the true data from noisy observations. The reverse transition \( p_{\theta}(\mathbf{x}_t | \mathbf{x}_{t+1}, \mathbf{H}, \mathbf{C}) \) is learned to denoise the data step by step.

The reverse process is defined as:
\[
p_{\theta}(\mathbf{x}_t | \mathbf{x}_{t+1}, \mathbf{H}, \mathbf{C}) = \mathcal{N}(\mathbf{x}_t; \mu_{\theta}(\mathbf{x}_{t+1}, \mathbf{H}, \mathbf{C}, t), \sigma_t^2 \mathbf{I}),
\]
where \( \mu_{\theta}(\mathbf{x}_{t+1}, \mathbf{H}, \mathbf{C}, t) \) is the predicted mean given by the neural network, conditioned on \( \mathbf{H} \), \( \mathbf{C} \), and the timestep \( t \), and \( \sigma_t^2 \) represents the fixed variance for each time step. The neural network is trained to predict the reverse dynamics of the diffusion process, progressively reducing noise over time while incorporating the conditional information \( \mathbf{H} \) and \( \mathbf{C} \).

In the conditional diffusion model, both historical app usage data \( \mathbf{H} \) and additional contextual features \( \mathbf{C} \) are explicitly incorporated into the diffusion process. These conditional inputs influence the noise schedule \( \beta_t \) and guide the denoising steps, enabling the model to generate app usage predictions that are tailored to specific historical and contextual scenarios.

We incorporate a loss function that ensures the predictions \( \hat{\mathbf{X}} \) are as accurate as possible. The loss function consists of two main components: the cosine similarity loss and the MSE loss.

The cosine similarity loss helps ensure that the predicted app usage matrix \( \hat{\mathbf{X}} \) and the true matrix \( \mathbf{X}_{\text{true}} \) are directionally similar. This is crucial for capturing the relative relationships between different app categories over time. The cosine similarity between two vectors \( \mathbf{a} \) and \( \mathbf{b} \) is defined as:
\[
\text{cosine\_similarity}(\mathbf{a}, \mathbf{b}) = \frac{\mathbf{a} \cdot \mathbf{b}}{\|\mathbf{a}\|_2 \|\mathbf{b}\|_2},
\]
and the cosine similarity loss for the entire matrix is calculated as:
\[
\mathcal{L}_{\text{cos}} = 1 - \frac{\sum_{t=1}^{T} \sum_{c=1}^{C} \hat{\mathbf{X}}_{t,c} \cdot \mathbf{X}_{\text{true}, t,c}}{\sqrt{\sum_{t=1}^{T} \sum_{c=1}^{C} \hat{\mathbf{X}}_{t,c}^2} \cdot \sqrt{\sum_{t=1}^{T} \sum_{c=1}^{C} \mathbf{X}_{\text{true}, t,c}^2}}.
\]

In addition to the cosine similarity, we apply the MSE loss to penalize pointwise errors between the predicted and true values. The MSE loss is expressed as:
\[
\mathcal{L}_{\text{MSE}} = \frac{1}{T \times C} \sum_{t=1}^{T} \sum_{c=1}^{C} (\hat{\mathbf{X}}_{t,c} - \mathbf{X}_{\text{true}, t,c})^2.
\]

The final loss function is a weighted sum of the MSE loss and cosine similarity loss:
\[
\mathcal{L} = \lambda_1 \cdot \mathcal{L}_{\text{MSE}} + \lambda_2 \cdot \mathcal{L}_{\text{cos}},
\]
where \( \lambda_1 \) and \( \lambda_2 \) are hyperparameters that control the balance between the two loss components. This combined loss ensures that the model both predicts the app usage values accurately and maintains the correct relative structure of the data.

The model is trained using gradient-based optimization methods such as Adam or SGD to minimize the loss function. During training, the neural network learns to map the noisy observations back to the original data distribution, progressively improving its ability to predict future app usage patterns based on historical data.

Once trained, the conditional diffusion model generates the predicted app usage matrix \( \hat{\mathbf{X}} \), which can be used for forecasting app usage over time across different categories. The model’s capacity to adapt to different conditioning information, including time-varying features, makes it highly effective in modeling app usage dynamics.

\subsection{2D Attention for APP Correlation}

\begin{figure}[t] 
    \centering
    \includegraphics[width=\columnwidth]{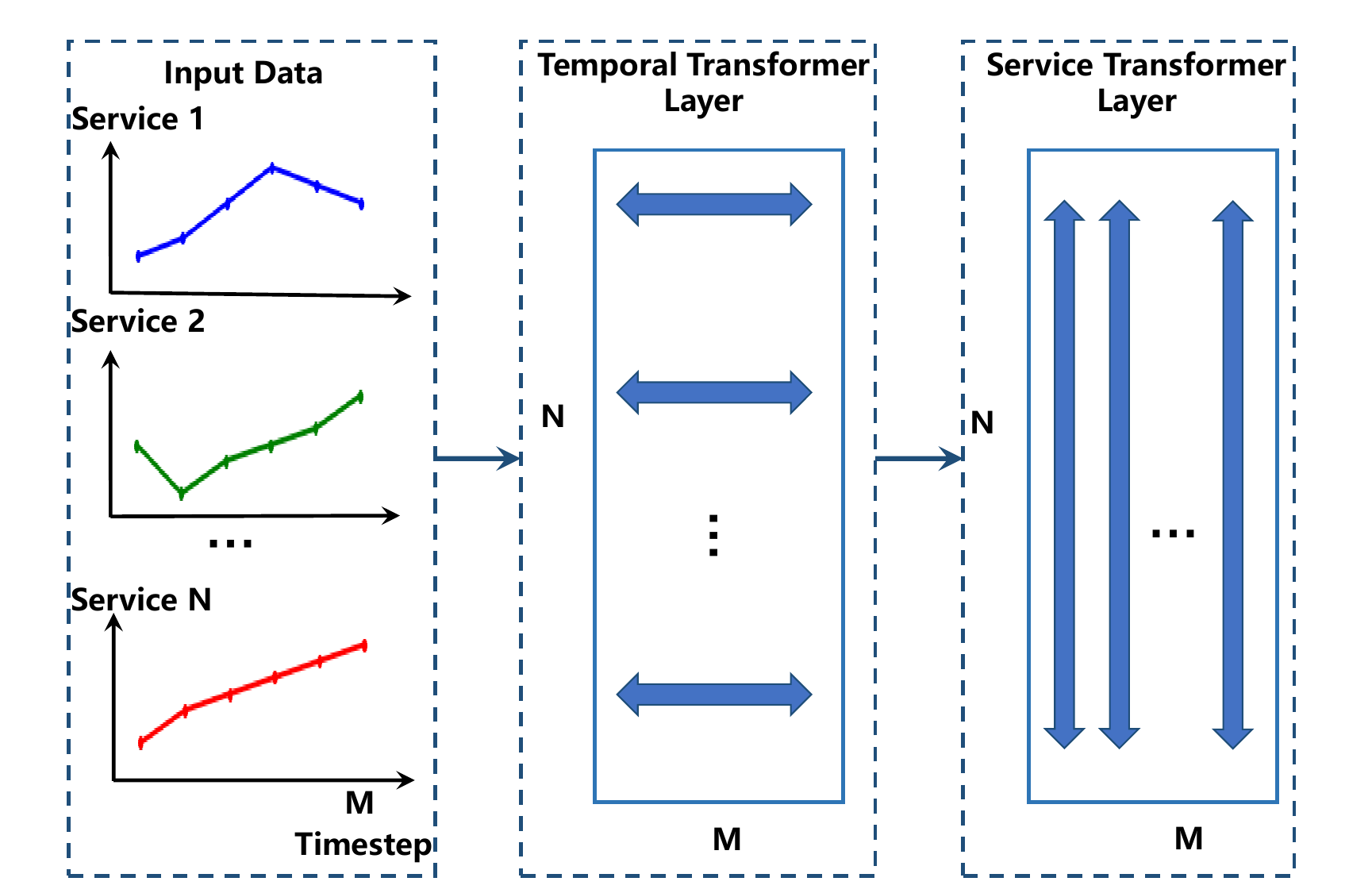} 
    \caption{The structure of 2D attention. The traffic corresponding to different services is first processed by a Transformer that handles the time dimension, and then by another Transformer that processes the dimensions representing different services.} 
    \label{fig_2D} 
\end{figure}

To capture the temporal and category correlations present in two-dimensional app usage data, we develop a two-layer transformer network in the denoiser. As shown in Figure~\ref{fig_2D}, this network utilizes a temporal transformer layer and a category transformer layer. The temporal transformer layer processes data for each app category to learn temporal dependencies, while the category transformer layer processes data for each time point to learn category dependencies. This dual-layer approach ensures a comprehensive understanding of temporal and category dynamics in the app usage data.

\begin{algorithm}[t]
\caption{Final Layer with 2D Transformer Structure}
\label{alg1}
\hspace*{0.02in} {\bf Input:}
Input tensor $x \in \mathbb{R}^{B \times C \times K \times L}$, conditioning vector $c \in \mathbb{R}^{B \times H}$, model parameters $\theta$\\
\hspace*{0.02in} {\bf Output:}
Output tensor $x_{\text{out}}$
{\small \begin{algorithmic}[1]
\State Compute the base shape of $x$: 
\State \qquad $(B, C, K, L) = \text{shape}(x)$
\State Reshape $x$ for the time dimension transformer: 
\State \qquad $x_{\text{time}} \leftarrow \text{reshape}(x, (B \cdot K, C, L))$
\State Pass $x_{\text{time}}$ through the time transformer: 
\State \qquad $x_{\text{time}} \leftarrow \text{TimeLayer}(x_{\text{time}})$
\State Reshape $x_{\text{time}}$ back: 
\State \qquad $x_{\text{time}} \leftarrow \text{reshape}(x_{\text{time}}, (B, C, K, L))$

\vspace{0.2cm}
\State Reshape $x$ for the feature dimension transformer: 
\State \qquad $x_{\text{feature}} \leftarrow \text{reshape}(x, (B \cdot L, C, K))$
\State Pass $x_{\text{feature}}$ through the feature transformer: 
\State \qquad $x_{\text{feature}} \leftarrow \text{FeatureLayer}(x_{\text{feature}})$
\State Reshape $x_{\text{feature}}$ back: 
\State \qquad $x_{\text{feature}} \leftarrow \text{reshape}(x_{\text{feature}}, (B, C, K, L))$

\vspace{0.2cm}
\State Apply LayerNorm and AdaLN modulation:
\State Compute shift and scale: 
\State \qquad $(\text{shift}, \text{scale}) \leftarrow \text{AdaLN}(c)$
\State Modulate $x$: 
\State \qquad $x \leftarrow \text{Norm}(x) \cdot \text{scale} + \text{shift}$

\State Apply the final linear transformation: $x_{\text{out}} \leftarrow \text{Linear}(x)$\;

\Return $x_{\text{out}}$\;
\end{algorithmic}}
\end{algorithm}

In the DiT architecture, a dual-layer Transformer structure is introduced to effectively process input tensors across both time and feature dimensions. The dual-Transformer design enables the model to simultaneously capture temporal dependencies and feature interactions, which enhances its ability to understand complex patterns in the data. The first Transformer layer captures lower-level temporal and feature information, while the second layer abstracts these details at a higher level, improving the model's representational power.
Additionally, Adaptive Layer Normalization (AdaLN) is employed to apply adaptive modulation to the output of the Transformer layers. AdaLN dynamically adjusts the model’s behavior based on the input data distribution, thus improving both stability and generalization. The specific implementation process is shown in Algorithm~\ref{alg1}. This adaptive mechanism is particularly useful for handling varying and complex input data.
Finally, a linear projection is applied to the transformed tensor, mapping the model’s internal representation to the target space for downstream tasks, such as classification or regression. This design choice ensures efficient information propagation while retaining the benefits of dual-dimensional processing and adaptive normalization.

\section{Performance Evaluation}
\label{per}
In this section, we present detailed experiment settings and overall prediction quality evaluations to validate our LSDM framework.
\subsection{Experiment Settings}

\subsubsection{Data collection}

This dataset contains anonymized records of user activities, including user identification, timestamps of HTTP requests or responses, packet lengths, visited domains, and user-agent fields. Apps are identified from network metadata using the SAMPLES framework~\cite{yao2015samples}, and their categories are determined by cross-referencing information from Android Market and Google Play.

Additionally, the dataset provides the spatial distribution of PoIs associated with each base station, enabling detailed analyses of user mobility patterns and network activity. As a comprehensive resource, this dataset supports research in mobile network traffic, urban analytics, and application-level behavior modeling, offering significant value for academic and industrial studies.

We retrieved corresponding satellite images based on the geographic coordinates of the base stations connected by users to provide richer multimodal environmental information.

\subsubsection{Comparative Methods}
We evaluate the performance of our proposed model against a variety of state-of-the-art baselines, spanning statistical methods, prompt-based approaches, transformer-based architectures, and advanced diffusion models, as exemplified by the baselines outlined below:  

\begin{itemize}
    \item \textbf{ARIMA}~\cite{xu2023machine}: Statistical methods integrating autoregression and moving averages to forecast time series.  
    \item \textbf{Tempo}~\cite{cao2023tempo}: Designs temporal prompts incorporating trend and seasonal features to enable time series forecasting using pre-trained models like GPT-2. 
    \item \textbf{Time-LLM}~\cite{jin2023time}: Leverages natural language descriptions of time series features as prompts for pre-trained language models like LLAMA-7B to make predictions.  
    \item \textbf{PatchTST}~\cite{nie2022time}: Decomposes time series into segments and employs transformers for feature extraction and prediction.  
    \item \textbf{NetDiffus}~\cite{sivaroopan2024netdiffus}: NetDiffus is an advanced method that employs time-series imaging and diffusion models for network data generation. 
    \item \textbf{DiT}~\cite{peebles2023scalable}: A novel diffusion model utilizes a transformer backbone to process latent image patches instead of the traditional U-Net architecture.      
    \item \textbf{CSDI}~\cite{tashiro2021csdi}: A conditional diffusion model incorporating 2D attention to capture complex dependencies in structured inputs for precise conditional generation.  
    \item \textbf{RF-diffusion}~\cite{chi2024rf}: A diffusion model based on the DiT architecture designed for the frequency domain, modified by us to operate effectively in the time domain. 
\end{itemize}

\subsubsection{Evaluation Metrics}
To evaluate the performance of our model comprehensively, we employ the following metrics, which measure various aspects of prediction accuracy and similarity:  
\begin{itemize}
    \item \textbf{Mean Squared Error (MSE):}
    MSE measures the average squared difference between predicted and true values. A smaller value indicates better performance.
    \begin{equation}
        \text{MSE} = \frac{1}{n} \sum_{i=1}^{n} (y_i - \hat{y}_i)^2.
    \end{equation}

    \item \textbf{Root Mean Squared Error (RMSE):}
    RMSE is the square root of MSE, providing an interpretable error metric in the same unit as the target variable.
     \begin{equation}
    \text{RMSE} = \sqrt{\frac{1}{n} \sum_{i=1}^{n} (y_i - \hat{y}_i)^2}.
    \end{equation}

    \item \textbf{Mean Absolute Error (MAE):}
    MAE measures the average magnitude of errors without considering their direction. A smaller value indicates better performance.
     \begin{equation}
    \text{MAE} = \frac{1}{n} \sum_{i=1}^{n} |y_i - \hat{y}_i|.
    \end{equation}

    \item \textbf{Cosine Similarity (CS):}
    CS evaluates the similarity between the predicted and true vectors by measuring the cosine of the angle between them.
    \begin{equation}
    \text{CS} = \frac{\sum_{i=1}^{n} y_i \hat{y}_i}{\sqrt{\sum_{i=1}^{n} y_i^2} \cdot \sqrt{\sum_{i=1}^{n} \hat{y}_i^2}}.
    \end{equation}

    \item \textbf{Coefficient of Determination ($R^2$):}
    $R^2$ represents the proportion of variance in the dependent variable that is predictable from the independent variable(s). Higher values indicate better performance.
     \begin{equation}
    R^2 = 1 - \frac{\sum_{i=1}^{n} (y_i - \hat{y}_i)^2}{\sum_{i=1}^{n} (y_i - \bar{y})^2},
     \end{equation}
    where $\bar{y}$ is the mean of the true values.
\end{itemize}

\subsection{Overall Prediction Quality}

The experimental results presented in Table~\ref{tab:comparison_models} demonstrate the effectiveness of various models in time series prediction tasks, evaluated across multiple metrics such as MSE, RMSE, MAE, CS, and $R^2$. Key observations are summarized below:  

\begin{figure*}[t] 
    \centering
    \includegraphics[width=\textwidth]{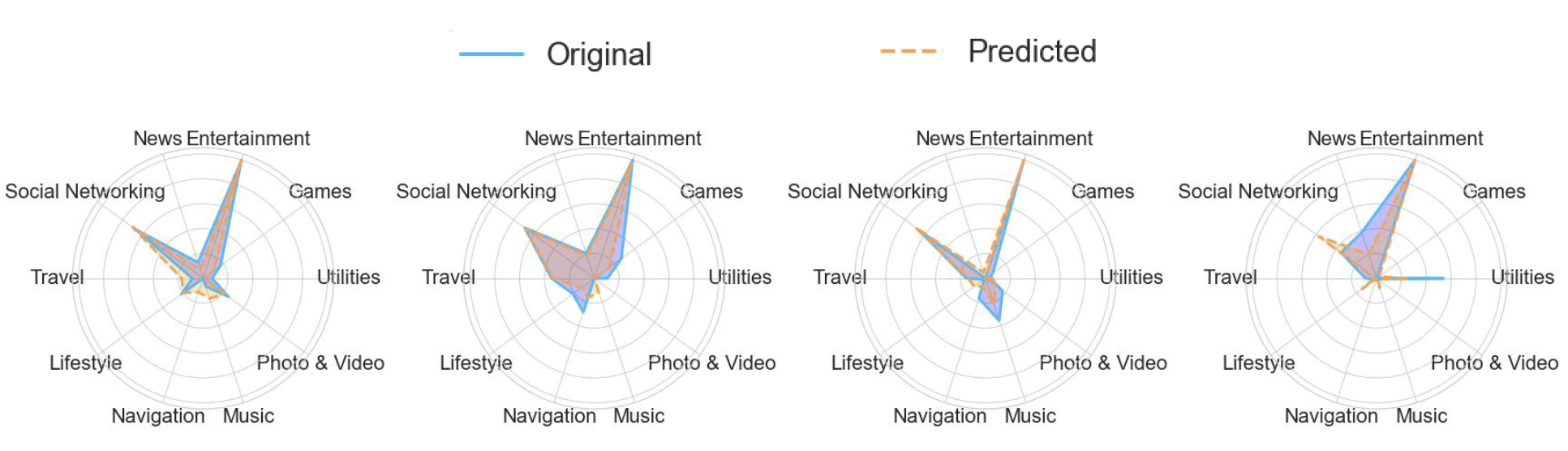} 
    \caption{User-level true vs. predicted comparison, we selected the prediction results of four groups of users with different usage preferences. In the radar chart, each set of lines represents the comparison between the real and predicted traffic for the corresponding services: the blue lines indicate the real values, while the orange lines represent the predicted values.} 
    \label{fig_cc} 
\end{figure*}

\begin{itemize}
    \item \textbf{Baseline Performance:} Traditional models like ARIMA show limited prediction accuracy with the highest MSE (1.015) and RMSE (1.008), highlighting the difficulty of handling complex temporal patterns using classical statistical methods.

    \item \textbf{Emerging Techniques:} Natural language-based models, such as Tempo and TimeLLM, achieve better performance by leveraging pre-trained large language models with temporal prompts. TimeLLM outperforms Tempo across all metrics, showcasing the benefits of advanced prompt design and larger model capacities.

    \item \textbf{Spatio-temporal Models:} PatchTST and NetDiffus further improve prediction accuracy by incorporating spatio-temporal correlations and advanced transformer-based architectures. PatchTST achieves a balance between accuracy and complexity, while NetDiffus reduces MSE to 0.0976, demonstrating superior performance in handling spatio-temporal dependencies.

    \item \textbf{Diffusion Models:} Models like DiT and RF-diffusion leverage diffusion mechanisms for robust time series predictions. Among these, DiT achieves one of the lowest MSEs (0.0892), indicating its effectiveness in capturing complex temporal patterns. However, RF-diffusion has a higher MAE, suggesting room for improvement in specific aspects.

    \item \textbf{Proposed Models:} The proposed models (\textit{Ours-Non condition} and \textit{Ours+condition}) achieve the best overall performance, with significant reductions in MSE (0.0719 and 0.0700) and RMSE (0.2287 and 0.2262). The inclusion of conditional information (\textit{Ours+condition}) further enhances performance, achieving the highest $R^2$ value (0.6639) and the lowest MAE (0.1264), confirming the effectiveness of the proposed approach in leveraging additional contextual information.
\end{itemize}

\begin{figure}[t] 
    \centering
    \includegraphics[width=1.0\columnwidth]{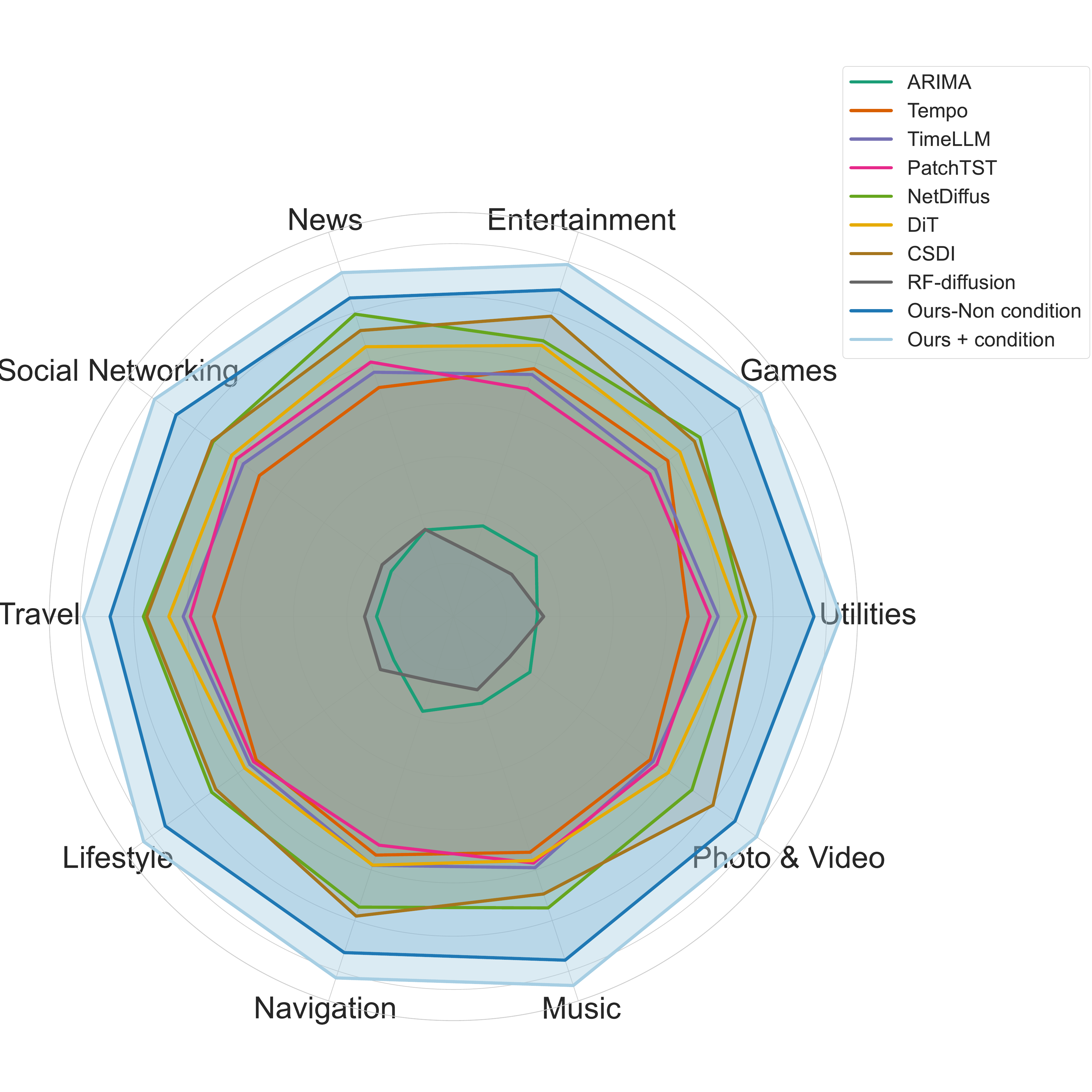} 
    \caption{Radar chart of prediction accuracy across app categories.} 
    \label{fig_comparison} 
\end{figure}

From the Figure~\ref{fig_comparison}, it is evident that our proposed model outperforms other baseline models in terms of prediction accuracy across all app categories. Specifically, our model demonstrates higher R² values for every app category, with notable improvements in categories such as \textit{Utilities}, \textit{Games}, and \textit{Music}, showcasing superior predictive performance compared to other models. Furthermore, when compared to the model without conditions, our conditional model shows enhanced accuracy in multiple categories, highlighting the effectiveness and robustness of our approach. As shown in Figure~\ref{fig_cc}, we compare the prediction results of four groups of users with different preferences and find that the model effectively captures their service usage preferences.

In conclusion, the proposed models set a new benchmark in time series forecasting, demonstrating their superiority over traditional statistical methods, natural language-based models, and advanced spatio-temporal architectures. The results validate the design choices and highlight the importance of incorporating conditional inputs for further improving prediction accuracy.

\begin{table*}[tb]
\centering
\large 
\vspace{-2mm}
\caption{Comparison of Metrics Across Models. \textbf{Bold} denotes the best results, and \underline{underline} denotes the second-best results. Column '$\Delta$' represents the improvement percentage relative to the 'Ours +condition' row. 'Non CON' represents not incorporating multimodal contextual information.}
\vspace{-2mm}
\resizebox{\textwidth}{!}{ 
\begin{tabular}{p{1.2in} | p{0.7in} p{0.7in} | p{0.7in} p{0.7in} | p{0.7in} p{0.7in} | p{0.7in} p{0.7in} | p{0.7in} p{0.7in}}
\toprule
\textbf{Model} & \textbf{MSE} & \boldmath{$\Delta_{MSE}$} & \textbf{RMSE} & \boldmath{$\Delta_{RMSE}$} & \textbf{MAE} & \boldmath{$\Delta_{MAE}$} & \textbf{CS} & \boldmath{$\Delta_{CS}$} & \boldmath{$R^2$} & \boldmath{$\Delta_{R^2}$} \\ 
\midrule
ARIMA & 1.015 & 1350.00\% & 1.008 & 346.02\% & 2.9342 & 2221.45\% & 0.3723 & 43.43\% & 0.1650 & 75.16\% \\ 
Tempo & 0.1953 & 179.00\% & 0.4417 & 95.19\% & 0.2115 & 67.39\% & 0.3542 & 46.15\% & 0.4693 & 29.35\% \\ 
TimeLLM & 0.1682 & 140.29\% & 0.4101 & 81.33\% & 0.1928 & 52.48\% & 0.3796 & 42.31\% & 0.4902 & 26.20\% \\ 
PatchTST & 0.1821 & 151.57\% & 0.4266 & 88.63\% & 0.2054 & 62.49\% & 0.3885 & 41.00\% & 0.4754 & 28.39\% \\ 
NetDiffus & 0.0976 & 39.43\% & 0.2726 & 21.43\% & 0.1826 & 44.45\% & 0.3401 & 48.33\% & 0.5723 & 14.00\% \\ 
DiT & 0.0892 & 27.43\% & 0.2474 & 9.37\% & 0.1922 & 52.05\% & 0.2936 & 55.39\% & 0.5089 & 23.35\% \\ 
CSDI & 0.0888 & 26.86\% & 0.2449 & 8.29\% & 0.1906 & 50.84\% & 0.3114 & 52.68\% & 0.5731 & 14.00\% \\ 
RF-diffusion & 0.0936 & 33.71\% & 0.2654 & 17.33\% & 0.1729 & 36.81\% & 0.5767 & 12.00\% & 0.1531 & 76.95\% \\ 

Ours & \textbf{0.0700} & 0 & \textbf{0.2262} & 0 & \textbf{0.1264} & 0 & \textbf{0.658} & 0 & \textbf{0.6639} & 0 \\ 
\bottomrule
\end{tabular}}
\label{tab:comparison_models}
\vspace{-1mm}
\end{table*}

\section{Analysis}
\label{ab}
In this section, we conduct ablation experiments to evaluate the contributions of different modules in LSDM. These experiments provide insights into the importance of each component and their impact on overall performance.

\subsection{Analysis of 2D-Attention}

The experimental results shown in Table~\ref{tab:comparison_2d_attention} illustrate the effectiveness of introducing 2D Attention. Compared to the baseline (i.e., Non 2D-Attention), our method demonstrates significant improvements across all evaluation metrics. In particular, metrics such as SSIM, CS, and $R^2$ exhibit notable gains, highlighting the ability of 2D Attention to effectively capture relationships among different apps. This structural enhancement enables the model to better represent complex interactions and dependencies, resulting in improved overall performance.

\begin{table*}[tb]
\centering
\large
\caption{Comparison of Metrics with and without 2D Attention. \textbf{Bold} denotes the best results, and \underline{underline} denotes the second-best results. Column '$\Delta$' represents the improvement percentage relative to the 'Ours' row.  'Non CON' represents not incorporating multimodal contextual information.}
\vspace{-2mm}
\resizebox{\textwidth}{!}{
\begin{tabular}{p{1.5in} | p{0.7in} p{0.7in} | p{0.7in} p{0.7in} | p{0.7in} p{0.7in} | p{0.7in} p{0.7in} | p{0.7in} p{0.7in}}
\toprule
\textbf{Method} & \textbf{MSE} & \boldmath{$\Delta_{MSE}$} & \textbf{RMSE} & \boldmath{$\Delta_{RMSE}$} & \textbf{MAE} & \boldmath{$\Delta_{MAE}$} & \textbf{CS} & \boldmath{$\Delta_{CS}$} & \boldmath{$R^2$} & \boldmath{$\Delta_{R^2}$} \\ 
\midrule
Non 2D-Attention & 0.0962 & 37.43\% & 0.2701 & 19.41\% & 0.1803 & 42.65\% & 0.4301 & 34.64\% & 0.0751 & 88.69\% \\ 
Directly adding CON & 0.0960 & 37.14\% & 0.2697 & 19.26\% & 0.1789 & 41.56\% & 0.5716 & 13.12\% & 0.0657 & 90.11\% \\ 
Ours-Non CON & \underline{0.0749} & 7.00\% & \underline{0.2392} & 5.76\% & \underline{0.1334} & 5.53\% & \underline{0.634} & 3.65\% & \underline{0.6452} & 2.83\% \\ 
Ours & \textbf{0.0700} & 0 & \textbf{0.2262} & 0 & \textbf{0.1264} & 0 & \textbf{0.6580} & 0 & \textbf{0.6639} & 0 \\ 
\bottomrule
\end{tabular}}
\label{tab:comparison_2d_attention}
\vspace{-1mm}
\end{table*}

\subsection{Analysis of Pretrained models}

In Table \ref{tabllm}, we compare five experimental setups designed to evaluate multimodal integration in models. The first setup, GPT-3~\cite{chatgpt}, processes image and text embeddings without fine-tuning for multimodal tasks, leading to the poorest performance (MSE = 0.0805) due to its limited ability to handle conflicting multimodal information. The second setup, GPT-4~\cite{achiam2023gpt}, shows improved performance (MSE = 0.0740), leveraging the inherent strength of GPT-4 but still lacking task-specific optimization. The third setup, GPT-4O~\cite{hurst2024gpt}, incorporates fine-tuning on multimodal tasks, yielding better results (MSE = 0.0725) and demonstrating the importance of tailored optimization for multimodal integration. The fourth setup, Pretrained CLIP, uses embeddings generated by a pretrained CLIP model, achieving moderate improvement (MSE = 0.0752) but not fully exploiting the potential of multimodal data. The final setup, Ours, fine-tunes the CLIP model to integrate multimodal data effectively, achieving the best results across all metrics, with the lowest MSE of 0.0700. This highlights the effectiveness of our approach in maximizing the complementary strengths of multimodal data, avoiding conflicts, and enhancing model performance for complex tasks.

The experimental results demonstrate the superiority of our proposed approach in effectively integrating multimodal information. While baseline methods such as GPT-3 and GPT-4 show limited performance improvements due to a lack of task-specific optimization, fine-tuned approaches like GPT-4O and pretrained CLIP partially address these issues but still fall short. Our fine-tuned CLIP model achieves the best results across all metrics. These findings highlight the importance of fine-tuning multimodal models to align the information from different modalities effectively, enabling complementary interactions and reducing information conflicts. This comprehensive improvement across metrics confirms that our approach is better suited for handling complex multimodal tasks and significantly outperforms existing methods.

\begin{table*}[tb]
\centering
\large
\caption{Comparison of Metrics Across Multimodal Models. \textbf{Bold} denotes the best results, and \underline{underline} denotes the second-best results. Column '$\Delta$' represents the improvement percentage relative to the 'Ours' row.}
\vspace{-2mm}
\resizebox{\textwidth}{!}{
\begin{tabular}{p{1.5in} | p{0.7in} p{0.7in} | p{0.7in} p{0.7in} | p{0.7in} p{0.7in} | p{0.7in} p{0.7in} | p{0.7in} p{0.7in}}
\toprule
\textbf{Method} & \textbf{MSE} & \boldmath{$\Delta_{MSE}$} & \textbf{RMSE} & \boldmath{$\Delta_{RMSE}$} & \textbf{MAE} & \boldmath{$\Delta_{MAE}$} & \textbf{CS} & \boldmath{$\Delta_{CS}$} & \boldmath{$R^2$} & \boldmath{$\Delta_{R^2}$} \\ 
\midrule
GPT-3  & 0.0805 & 15.00\% & 0.2438 & 7.79\% & 0.1348 & 6.65\% & 0.6230 & 5.32\% & 0.6090 & 8.25\% \\ 
GPT-4  & 0.0740 & 5.71\% & 0.2305 & 1.90\% & 0.1287 & 1.82\% & 0.6500 & 1.22\% & 0.6450 & 2.85\% \\ 
GPT-4O  & \underline{0.0725} & 3.57\% & \underline{0.2280} & 0.80\% & \underline{0.1273} & 0.71\% & \underline{0.6550} & 0.46\% & \underline{0.6580} & 0.89\% \\ 
Pretrained CLIP  & 0.0752 & 7.43\% & 0.2345 & 3.67\% & 0.1305 & 3.25\% & 0.6470 & 1.67\% & 0.6350 & 4.34\% \\ 
Ours & \textbf{0.0700} & 0 & \textbf{0.2262} & 0 & \textbf{0.1264} & 0 & \textbf{0.6580} & 0 & \textbf{0.6639} & 0 \\ 
\bottomrule
\end{tabular}}
\label{tabllm}
\vspace{-1mm}
\end{table*}

\begin{figure}[t] 
    \centering
    \includegraphics[width=\linewidth]{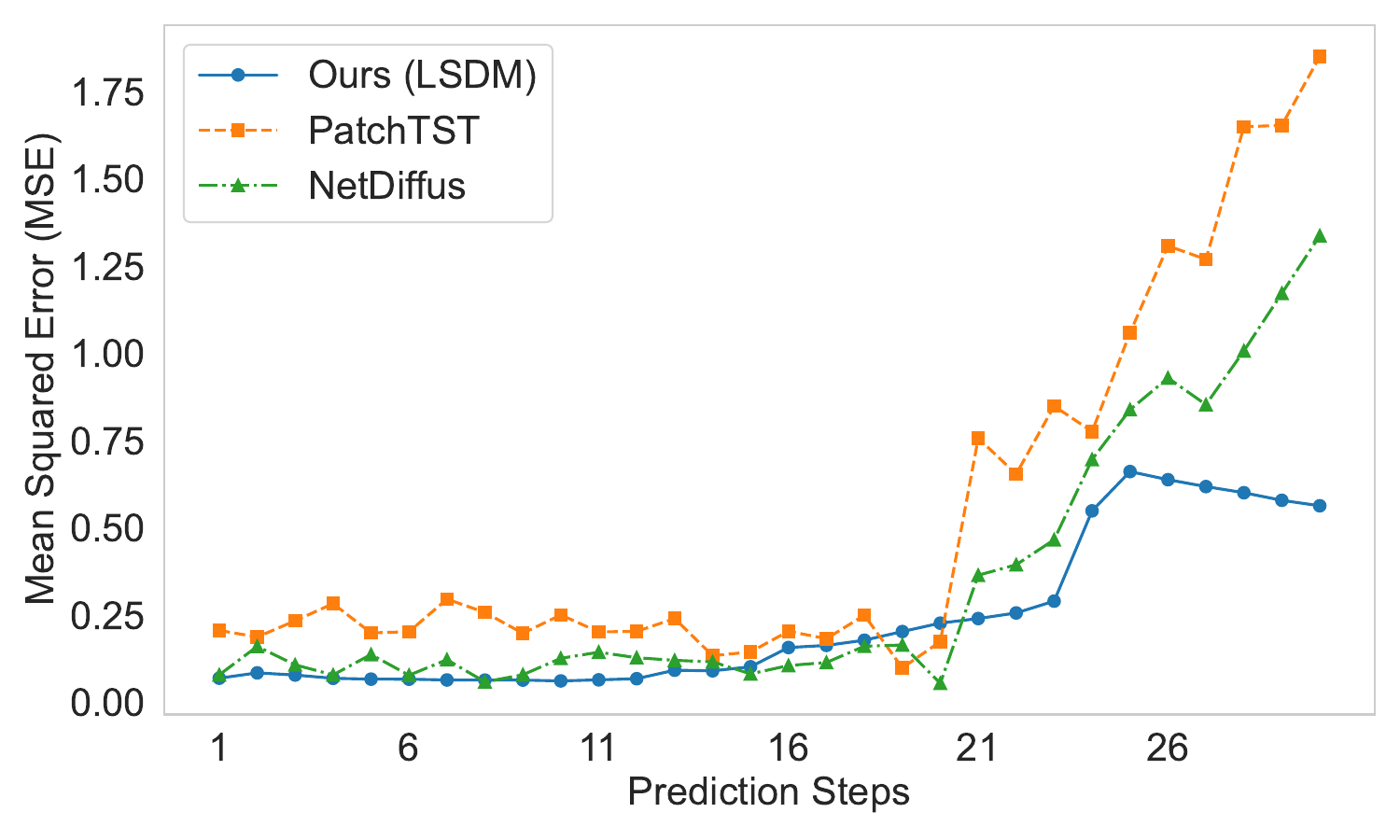} 
    \caption{Comparison of MSE trends across prediction steps for different models.} 
    \label{fig_line} 
\end{figure}

\subsection{Scalability}

\subsubsection{Experimental Objectives}
The primary objective of this experiment is to evaluate the stability and performance of the proposed model in predicting user service traffic over extended time steps. By sequentially predicting multiple future time steps, the experiment aims to verify the model's ability to capture user preferences and traffic trends accurately.

\subsubsection{Experimental Setup}
\begin{itemize}
    \item \textbf{Prediction Range:} The experiment involves predicting over 10, 20, and 30 time steps sequentially, representing short-term, mid-term, and long-term prediction tasks.
    \item \textbf{Prediction Method:} At each time step, the predicted value is used as a condition for predicting the next time step, simulating real-world dynamic scenarios through recursive prediction.
    \item \textbf{Evaluation Metrics:} MSE is used to measure the proximity between predicted and true values. Additionally, heatmaps are used for comparative analysis of the time-service distribution of predictions and ground truth.
\end{itemize}

\subsubsection{Experimental Procedure}
\begin{enumerate}
    \item \textbf{Sequential Prediction:} Starting from the initial time step, the model predicts the first step based on true data, and each subsequent prediction uses the previous step's predicted value as input, iterating until completing the 10, 20, or 30-step prediction tasks.
    \item \textbf{Post-Processing:} To maintain the stability and alignment of predictions with true values, noise suppression and normalization techniques are applied to the predicted results.
    \item \textbf{Visualization:} Heatmaps are generated to visualize and compare the predicted and true values across different prediction lengths.
\end{enumerate}

\subsubsection{Model Stability Analysis}

\begin{figure*}[tb] 
    \centering
    \includegraphics[width=0.9\textwidth]{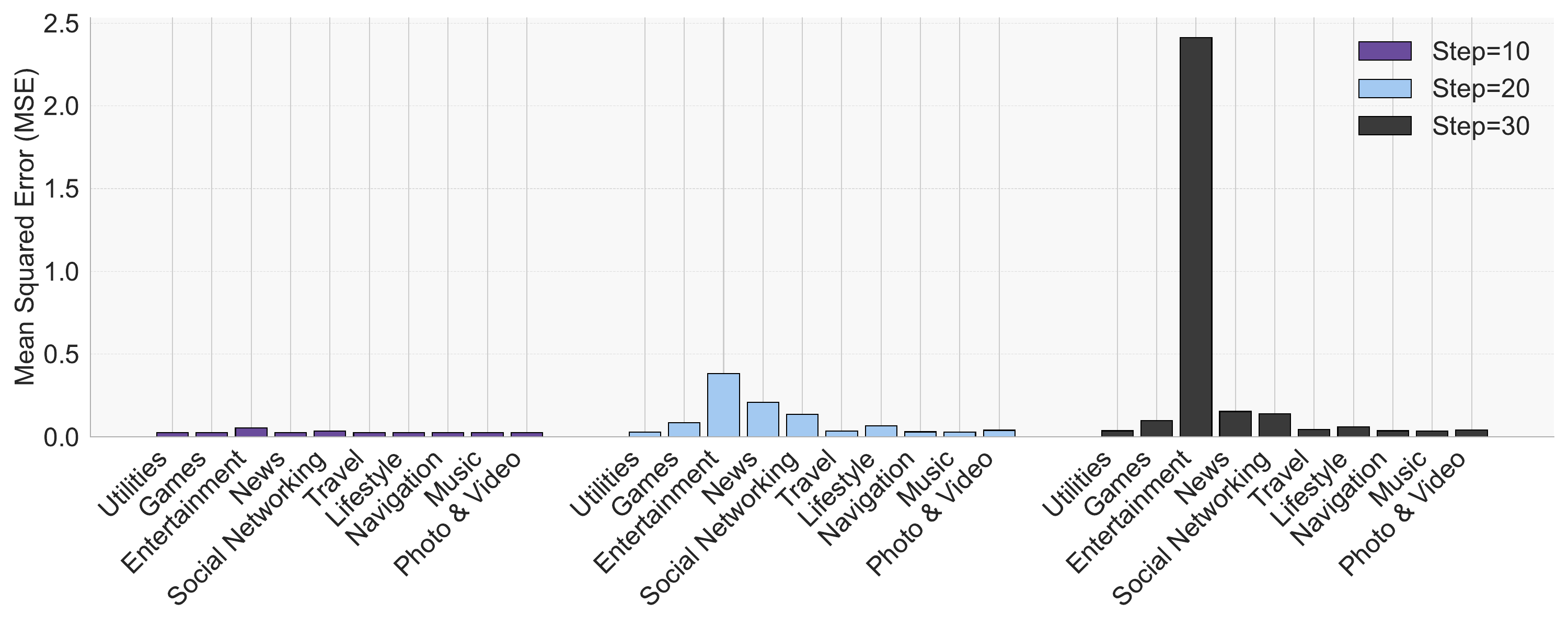} 
    \caption{MSE comparison for different services grouped by prediction steps.} 
    \label{fig_h} 
\end{figure*}

From Figure \ref{fig_line}, it is evident that our model demonstrates stable and excellent performance in the first 20 prediction steps, maintaining a consistently low MSE. However, at step 23, the MSE suddenly increases, indicating a brief instability. This issue is mitigated by incorporating environmental information as conditions into the large model’s background knowledge, resulting in a gradual reduction of MSE in subsequent prediction steps. In contrast, models like PatchTST and NeiDiffus exhibit significant and fluctuating increases in MSE during later prediction steps. This instability stems from their insufficient ability to capture the diverse predictive characteristics of app categories, leading to error accumulation and degraded performance. Additionally, our model demonstrates robust stability in recursive prediction tasks, where errors do not accumulate significantly for prediction lengths of 10, 20, or even 30 steps. This highlights the model’s capability to effectively mitigate error propagation, ensuring reliable performance even in longer prediction scenarios.

Figure \ref{fig_h} further illustrates the MSE trends of different service categories under varying prediction step lengths. For most service categories, MSE gradually increases as the prediction steps extend, reflecting the inherent challenges of long-term prediction. However, certain service categories, particularly entertainment-related ones, are highly sensitive to the prediction step length. These services exhibit significant traffic fluctuations, often accompanied by pronounced peaks, with peak times and magnitudes varying across different users. This dynamic nature makes accurate predictions for such services increasingly difficult as the prediction steps grow, resulting in a sharp decline in prediction accuracy.

In summary, our model exhibits superior performance and stability in long-term prediction tasks, effectively mitigating error propagation and outperforming other baseline methods. While service categories with significant fluctuations remain challenging, incorporating environmental information as conditions provides a promising solution to enhance prediction performance in these scenarios.

\subsubsection{Traffic Distribution Capturing}
\begin{itemize}
    \item \textbf{Short-term Prediction (10 Steps):} The predicted traffic distribution closely aligns with the actual user traffic patterns, accurately reflecting the trends in traffic changes.
    \item \textbf{Mid-term Prediction (20 Steps):} The overall traffic distribution remains highly consistent with the true values, with only minor deviations observed in certain time intervals, which do not affect the accurate capture of overall trends.
    \item \textbf{Long-term Prediction (30 Steps):} Over longer time horizons, the model successfully captures the primary traffic distribution trends. While slight amplification of fluctuations is observed, the overall traffic dynamics remain in agreement with the actual patterns.
\end{itemize}

\section{Conclusion}
In this paper, we investigated the problem of context-aware multimodal app usage and traffic prediction. To address this challenge, we propose LSDM, a novel prediction framework that combines three key innovations. First, LSDM integrates environmental contextual information through LLM-based background knowledge. Second, it employs a conditional diffusion model as its primary architecture for precise prediction. Third, it incorporates a two-dimensional attention mechanism to effectively capture traffic correlations between different apps. Extensive experiments on real-world datasets demonstrate the robustness and accuracy of our approach, compared to advanced diffusion models, the \(R^2\) improves by at least 14\%, and in multi-step predictions, the MSE remains below 0.75 for up to 30 steps. Future work will focus on optimizing the framework for real-time app traffic prediction, enhancing multimodal context integration, and supporting adaptive network resource management in mobile environments.


\bibliographystyle{IEEEtran}
\bibliography{references}

\end{document}